\documentclass[a4paper,conference]{IEEEtran}

\usepackage{graphicx}
\usepackage{amsmath}
\usepackage{amssymb}
\usepackage{paralist}
\usepackage[T1]{fontenc}
\usepackage[hyphens]{url}
\usepackage[breaklinks=true,
  colorlinks=false,
  pdfborder={0 0 0},
  pageanchor=false,
  bookmarks=false]{hyperref}

\begin{document}

\title{An end-to-end Optical Character Recognition approach for
  ultra-low-resolution printed text images}

\author{\IEEEauthorblockN{Julian Gilbey}
  \IEEEauthorblockA{Department of Applied Mathematics and Theoretical\\
    Physics,
    University of Cambridge\\
    Email: \href{mailto:jdg18@cam.ac.uk}{jdg18@cam.ac.uk}\\
    ORCID iD: \href{https://orcid.org/0000-0002-5987-5261}{0000-0002-5987-5261}
  }
  \and
  \IEEEauthorblockN{Carola-Bibiane Sch\"onlieb}
  \IEEEauthorblockA{Department of Applied Mathematics and Theoretical\\
    Physics,
    University of Cambridge\\
    Email: \href{mailto:cbs31@cam.ac.uk}{cbs31@cam.ac.uk}\\
    ORCID iD: \href{https://orcid.org/0000-0003-0099-6306}{0000-0003-0099-6306}
  }
}

\maketitle

\begin{abstract}
  Some historical and more recent printed documents have been scanned
  or stored at very low resolutions, such as 60 dpi.  Though such
  scans are relatively easy for humans to read, they still present
  significant challenges for optical character recognition (OCR)
  systems.  The current state-of-the art is to use super-resolution to
  reconstruct an approximation of the original high-resolution image
  and to feed this into a standard OCR system.  Our novel end-to-end
  method bypasses the super-resolution step and produces better OCR
  results.  This approach is inspired from our understanding of the
  human visual system, and builds on established neural networks for
  performing OCR.

  Our experiments have shown that it is possible to perform OCR on
  60~dpi scanned images of English text, which is a significantly
  lower resolution than the state-of-the-art, and we achieved a mean
  character level accuracy (CLA) of 99.7\% and word level accuracy
  (WLA) of 98.9\% across a set of about 1000 pages of 60~dpi text in a
  wide range of fonts.  For 75~dpi images, the mean CLA was 99.9\% and
  the mean WLA was 99.4\% on the same sample of texts.  We make our
  code and data (including a set of low-resolution images with their
  ground truths) publicly available as a benchmark for future work in
  this field.
\end{abstract}

\section{Introduction}

The current generation of optical character recognition (OCR) software
is designed for recognising printed text at 300~dpi.  For lower
resolutions down to about 150~dpi, the image can be enlarged to
300~dpi prior to performing the OCR step using bicubic interpolation,
and the results are still excellent.  (For example, when testing this
on our scanned text dataset, described in section~\ref{sec:data}, we
found that on average, reducing the 300~dpi images by a factor of~2
and then enlarging them back to 300~dpi using bicubic interpolation
resulted in slightly \emph{better} recognition results when using the
standard Tesseract OCR software.  An analysis of this behaviour is
beyond the scope of this paper.)

Document images at resolutions less than about 150~dpi offer much
greater challenges, as much less information is present in the image.
(See, for example \cite{dong_boosting_2015},
\cite{habeeb_improving_2014} and the other sources discussed in
section~\ref{sec:literature}.  We give illustrative numerical results
for our dataset in section~\ref{sec:results} below.)  This might
happen if images are stored at a very low resolution for space
purposes, or if photographs of scenes containing text have been
captured from a distance so the text within them is very small.  In
this work, we have focused our attention on scanned documents, but
there are clear potential extensions of our technique to other use
cases.

\begin{figure}[t]
  \begin{center}
    \includegraphics[width=0.9\linewidth]{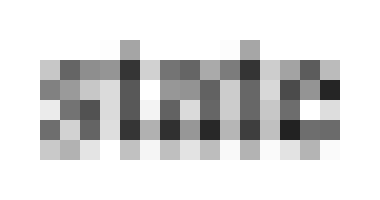}
  \end{center}
  \caption{A word from a 60 dpi scan, enlarged so that the individual
    pixels are visible.}
  \label{fig:example1}
\end{figure}

\begin{figure}[t]
  \begin{center}
    \includegraphics[width=0.9\linewidth]{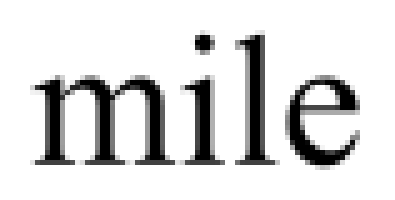}
  \end{center}
  \caption{A simulated word from a 300 dpi scan, enlarged so that the
    individual pixels are visible.}
  \label{fig:example2}
\end{figure}

It is astonishing that humans are able to read such low-resolution
images with relative ease when printed at this low resolution.
However, things are very different when they are enlarged.  For
example, figure~\ref{fig:example1} shows a word in a 60~dpi document,
enlarged so that the individual pixels are visible.  It is difficult
to visually determine what the word is and even where the character
boundaries are.  In spite of this, when the image is held at a
distance (or made smaller), the characters become quite distinct and
reasonably easy to read.  In contrast to this,
figure~\ref{fig:example2} shows a different word at 300~dpi, enlarged
to the same size; this is easily readable from both nearby and from a
distance.  A standard OCR system trained to recognise 300~dpi images
of 10~point text will have significant difficulty with
figure~\ref{fig:example1}, even if it is enlarged to 300~dpi using
bicubic interpolation (as in Figure~\ref{fig:example1filtbc}).

In this work, we have taken a novel approach to this problem, building
on certain existing ideas, but introducing a new human-vision-inspired
technique.  Our main contributions are:
\begin{itemize}
\item Performing OCR on 60~dpi scanned printed text, and achieving
  outstanding character and word recognition rates; OCR on such
  ultra-low-resolution images has not been previously reported.
\item Performing OCR on 75~dpi images, and achieving character and
  word recognition rates significantly in excess of previously
  reported results for some fonts.
\item Demonstrating how sensitive OCR of low-resolution images can be
  to the font used, suggesting that it might be very hard to develop
  an OCR system which works well on all low-resolution images.
\end{itemize}

We note that we cannot directly compare our results with those
appearing in the literature, as there is no standard benchmark dataset
for this problem and none of the software developed is publicly
available.  We are therefore limited to quoting the results appearing
in those papers.  Also, as we note later, low-resolution OCR is very
sensitive to the font used and to the precise nature of the images,
making it even harder to directly compare our results.  To help remedy
this situation, we have prepared a test dataset that could be used for
future benchmarking, as described in section~\ref{sec:data} below.

The rest of the paper is structured as follows.  In
section~\ref{sec:literature}, we briefly survey related work.  In
section~\ref{sec:approach}, we discuss one aspect of human vision and
how it is related to our problem, and then explain our approach in
detail.  We present and analyse our results in
section~\ref{sec:results}.  A summary of our algorithm is displayed in
figures~\ref{fig:pipeline} to~\ref{fig:merging}.

\section{Related work}
\label{sec:literature}

One early approach for performing OCR on low-resolution text images
was presented in \cite{jacobs_low_2009} and \cite{jacobs_text_2005} in
the context of images captured by low-resolution cameras.  The issues
they faced were primarily around blurriness; the image resolution was
on the effective order of 150--200~dpi.  They used a neural network to
recognize individual characters, followed by a separate word
recognition phase.

More recently, there has been renewed effort to address low-resolution
images in the 75--100~dpi range.  Several authors consider using
multiple thresholds for binarisation of the images and then using an
ensemble method on the OCR output produced from each (such as
\cite{habeeb_improving_2014}, \cite{habeeb_enhanced_2018} and
\cite{lund_ensemble_2014}); these approaches are suitable for OCR
systems that work by initially thresholding a greyscale image to
produce a binary image before performing segmentation and feature
extraction.

A second popular approach has been to use super-resolution (for
example, \cite{ma_super_2013}, \cite{dong_boosting_2015},
\cite{pandey_new_2017}, \cite{pandey_efficient_2018}
and~\cite{lat_enhancing_2018}).  This approach was also perhaps
partially spurred on by the ICDAR 2015 Competition on Text Image
Super-Resolution (\cite{peyrard_icdar2015_2015},
\cite{icdar_icdar_2015} and \cite{peyrard_icdar2015-textsr_2020}),
which aimed to recognise text from frames extracted from video
streams.  The task was to reconstruct a high-resolution image from a
low resolution one, and to then use that to perform OCR.  (It is
important to note that super-resolution is not simply upscaling using
bicubic interpolation or similar, but rather an attempt to match the
original image.  The quality of a super-resolution reconstruction is
typically measured using a metric such as PSNR, comparing the
reconstruction with the ground truth high-resolution image.)  There
has been considerable success with this method.  For example, Pandey
et al.~\cite{pandey_efficient_2018} reported OCR results for 75~dpi
English text as 75.1\% CLA (character level accuracy) and 48\% WLA
(word level accuracy).  They achieved significantly better results on
75~dpi Tamil (90.8\% CLA and 54.3\% WLA) and Kannada (95.16\% WLA and
70.80\% WLA).  A more powerful super-resolution technique used by Lat
and Jawahar~\cite{lat_enhancing_2018}, and their quoted results are
the best we are aware of.  Their OCR results depended upon the OCR
tool being used: the commercial ABBYY achieved 97.88\% CLA and 95.20\%
WLA on their \textsc{swd} dataset, while the open source Tesseract
version 3.x achieved 85.96\% CLA and 89.70\% WLA; ABBYY also achieved
99.65\% CLA and 98.90\% WLA on their \textsc{end} dataset.

There are some issues with Lat and Jawahar's~\cite{lat_enhancing_2018}
methodology, which makes it difficult to directly compare their
results with other quoted results.  Though the precise texts chosen
for their \textsc{end} test set are different from the training texts,
the generation process was presumably the same, and so this only shows
(the still impressive result) that their GAN super-resolution scheme
can very accurately reconstruct high-resolution images from
low-resolution images if they are of a very similar nature to the
training images.  However, with real-life scanners, different scanners
may produce different behaviours from simulations, and so it is
necessary to test any proposed system with real scanners.  Also, for
their \textsc{swd} dataset, it seems from their Figure~7 that these
images are captured at closer to 100~dpi than 75~dpi, so it is
unsurprising that they give relatively good results.

None of these researchers reported attempting OCR with 60~dpi images:
whilst even 50~dpi images were improved using super-resolution (for
example, Pandey and Ramakrishnan \cite{pandey_efficient_2018} report
an average PSNR of 16.89 for super-resolved 50~dpi images and 19.36
for 75~dpi images), the OCR systems are relatively poor at extracting
text from such noisy images, and so the accuracy rate would be even
lower for super-resolved 60~dpi images.  We also note at this point
that 60~dpi images contain only about 60\% as many pixels as the
corresponding 75~dpi images, making the task significantly harder.

Super-resolving the image is of interest to humans, who would like to
see a higher-resolution image, but our primary goal here is to extract
text from the image.  Furthermore, as is clear from
figure~\ref{fig:example1} above, it can be challenging to even
determine whether such a low-resolution image is even of a sans serif
or a serif font.  A super-resolved 300~dpi image starting from this
would be quite likely to hallucinate much of the detail, and even if
it did perform this step reasonably well, a standard OCR system might
well struggle with the resulting image.

One possible approach is to create an end-to-end loss function, where
a single neural network performs super-resolution followed by OCR, and
the loss function is determined by the OCR accuracy.  Another
approach, which is close to the one we have taken in this paper, is to
skip the super-resolution step entirely, and to perform OCR directly
on the original image.  Such an approach was, to the best of our
knowledge, first taken by Wang and Singh~\cite{wang_systems_2018}, who
designed a RNN-LSTM neural network to replace Tesseract 3.02's
feature-based character recognition system, and trained a network on a
collection of domain-specific text lines generated at 72~dpi, 100~dpi,
150~dpi, 200~dpi and 300~dpi.  This RNN-LSTM network had an input
height of 32 pixels, so they scaled the images of text lines, both
during training and recognition, to this height using spline
interpolation.  (We emphasise that this scaling is not an attempt to
perform super-resolution.)  Their quoted result for 72~dpi images is a
CLA of 91.95\%.  Since then, Tesseract 4.0.0 has been released, which
uses a similar neural network for recognition, but is only trained on
300~dpi images.

\section{Our human-vision inspired approach}
\label{sec:approach}

\subsection{Human vision}
\label{ssec:humanvision}

\begin{figure}[t]
  \begin{center}
    \includegraphics[width=0.9\linewidth]{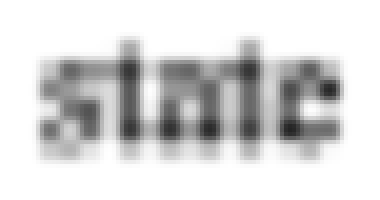}
  \end{center}
  \caption{Figure \ref{fig:example1}, enlarged to 300 dpi with
    nearest-neighbour interpolation and then filtered with a low-pass
    spatial Gaussian filter.}
  \label{fig:example1filt}
\end{figure}

\begin{figure}[t]
  \begin{center}
    \includegraphics[width=0.9\linewidth]{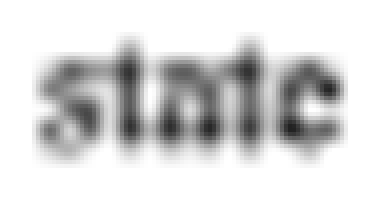}
  \end{center}
  \caption{Figure \ref{fig:example1}, enlarged to 300 dpi using
    bicubic interpolation.}
  \label{fig:example1filtbc}
\end{figure}

It is remarkable that most people are able to read the word in
figure~\ref{fig:example1} when they hold it at a distance.  The cause
of this was researched in detail by Majaj
et~al.~\cite{majaj_role_2002}.  At a far distance (or at a small size,
which is effectively the same), our visual system primarily processes
the low-frequency information (the gross shape of the letters),
whereas at a near distance (or large size), we process
higher-frequency information (the edges of the letters).  For a
low-resolution text image such as this, the high-frequency (edge)
information is mostly noise, so it is impossible to identify letters
from this, whereas the low-frequency information still contains
approximately enough information to identify the letters.  (The
concept that different frequencies are perceived at different
distances has also been successfully developed by Oliva et
al~\cite{oliva_hybrid_2006} to create hybrid images, for example
making a piece of text visible from nearby but invisible at a
distance; a well-known example of this phenomenon is the Marilyn
Monroe and Albert Einstein hybrid.)

One way we could therefore hope to improve the performance of an OCR
system on these low-resolution images is to upscale them to 300~dpi
(the resolution on which the OCR systems are trained) using
nearest-neighbour interpolation and then blur the images using a small
Gaussian filter.  This has the effect of reducing the high-frequency
components and making the edges less distinct.  A similar effect
results from using bilinear or bicubic interpolation.  As a result,
the word becomes easier to read; results of doing this to the word in
figure~\ref{fig:example1} are shown in figure~\ref{fig:example1filt}
and figure~\ref{fig:example1filtbc}.

\subsection{Current OCR systems}
\label{ssec:currentocr}

\begin{figure*}[t]
  \begin{center}
    \includegraphics[width=0.95\linewidth]{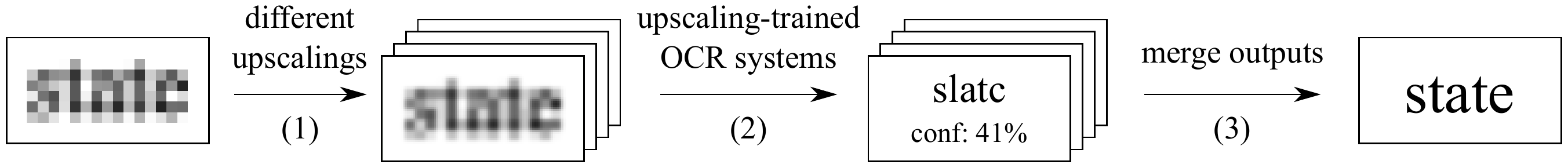}
  \end{center}
  \caption{A schematic illustration of our low-resolution OCR
    ensembling approach: (1) a low-resolution image is upscaled and
    blurred in multiple different ways; (2) each upscaled image is
    recognised using an LSTM trained for that type of upscaling, the
    results of which include confidences for each word; (3)~the
    results are merged to produce the best option for each word,
    taking a dictionary into account.  An example of the merging
    algorithm in action is shown in figure~\ref{fig:merging}.}
  \label{fig:pipeline}
\end{figure*}

\begin{figure*}[t]
  \begin{center}
    \includegraphics[width=0.95\linewidth]{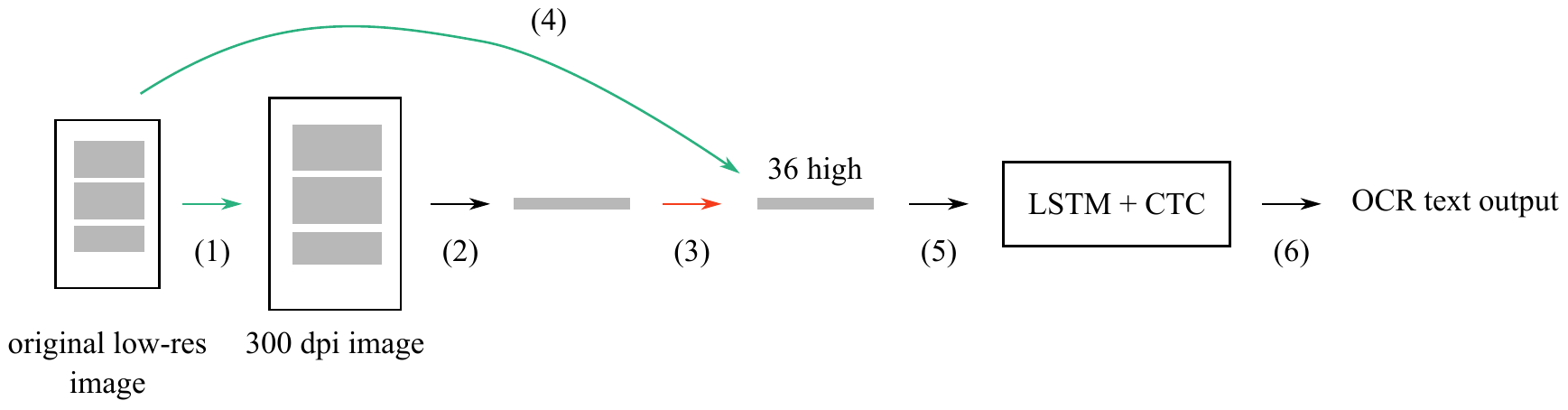}
  \end{center}
  \caption{A schematic illustration of our modified version of
    Tesseract for low-resolution images, where the green arrows
    indicate new steps and the red arrow indicates a replaced step:
    (1) the original low-resolution image is upscaled and blurred to
    make a 300~dpi image; (2)~the page is segmented and the text lines
    extracted; (3)~(original Tesseract only) each text line is scaled
    to 36 pixels high; (4)~each 36-pixel high scaled text line is
    produced using upscaling and blurring directly from the original
    image; (5)~the text line is passed to an LSTM network with a CTC;
    (6)~textual output and confidences for each word are produced.}
  \label{fig:tesseract}
\end{figure*}

\begin{figure*}[t]
  \begin{center}
    \includegraphics[width=0.95\linewidth]{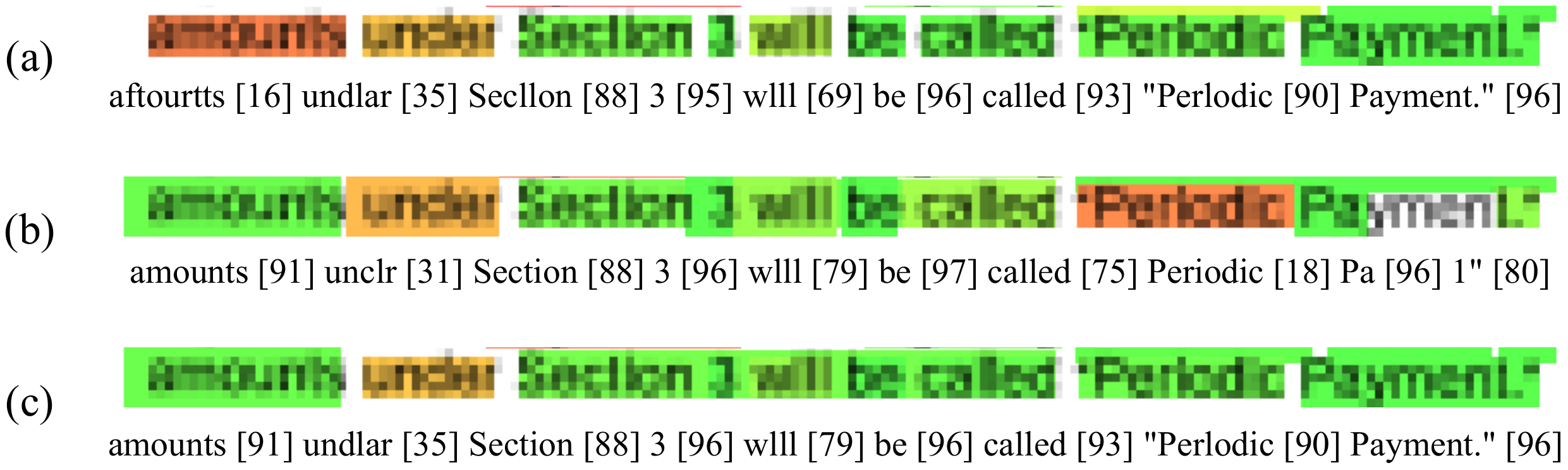}
  \end{center}
  \caption{Merging different results, the final step of the pipeline
    in figure~\ref{fig:pipeline}.  (a)~The words identified when
    performing OCR using one type of upscaling; the colours are a
    visual representation of the confidence level, and the textual
    results are shown below it (with a number such as [16] meaning
    16\% confidence).  (b)~The results using a different upscaling.
    (c)~Merger of (a) and~(b) using our algorithm.  The small bands
    of colour above and below the images are from adjacent lines of
    text.  Note that the bounding boxes of the identified words
    differ significantly between the two images.}
  \label{fig:merging}
\end{figure*}

State-of-the-art OCR systems now use trained bidirectional-LSTM neural
networks to perform the character recognition step.  Probably the
best-known open source OCR system available today is
Tesseract~\cite{smith_tesseract_nodate}, which has used such a network
since version 4.0.0 (see, for example, \cite{graves_offline_2009},
section~6 of~\cite{smith_tesseract_2016} and
\cite{ul-hasan_generic_2016}).  In brief, Tesseract first segments the
page into areas of text, which are further subdivided into paragraphs
and then lines.  Each line is passed into the trained network, which
uses a final softmax layer attached to a CTC (connectionist temporal
classifier) (\cite{graves_connectionist_2006},
\cite{smith_tesseract_2016}) to determine the most likely sequence of
letters and spaces; these are then further processed to produce a
sequence of words.  Each letter and each word also carries with it a
confidence score, based on the output of the classifier, and Tesseract
can be instructed to output this information.

\subsection{Modifying Tesseract}

Tesseract is designed to be trainable, and so we would like to train
it to recognise upscaled images.  A small issue arises in training it
directly on upscaled images, so we slightly modified the program to
handle them, as we now explain.  A sketch of Tesseract's pipeline
(version 4.1.1 of the software) is shown in
figure~\ref{fig:tesseract}, together with our modifications.  We
choose an upscaling method and apply it consistently for training and
recognition.  The upscalings that we used in this work were: upscaling
using nearest neighbour interpolation followed by a spatial Gaussian
filter with standard deviation 0, 0.5, 1, 1.5 or 2 pixels in the
300~dpi upscaled image; bilinear interpolation (with no filtering),
and bicubic interpolation (also with no filtering).  (In our
implementation, we have referred to nearest neighbour interpolation as
box interpolation; for upscaling, these are equivalent.)  Tesseract is
given the low-resolution image and upscales it to 300~dpi in the
manner specified.  It then performs its page segmentation as normal,
and extracts a text line.  At this point, Tesseract would scale the
text line to make it the correct height for the trained LSTM network
(which happens to be 36~pixels), using linear interpolation and
unsharp masking.  Since our image has already been upscaled once, this
would potentially lead to the loss of information, so we instead
produce the required region by upscaling the original image.  (This is
why we chose to modify the software rather than directly training on
upscaled images.)  During training, we use this image to train the
LSTM network for this type of upscaling, and during recognition, we
use that specifically-trained network.

For the training phase, we used transfer learning, beginning with the
Tesseract `best' English network (available from the Tesseract
repository).  We generated lines of low-resolution text in a variety
of fonts at a variety of font sizes (from 9pt to 12pt), adding small
amounts of Gaussian noise to the results, and also randomly rotating
half of them by a small angle (normally distributed with mean
$0^\circ$ and standard deviation $0.5^\circ$) to mimic the effect of
scans being slightly askew.  If we were training for a specific type
of document, it would make sense to train using lines of text from
that genre of document, as the LSTM will learn patterns of languages.
(This is what is done in~\cite{wang_systems_2018}.)  However, we have
chosen here to train our system for a wide variety of texts, so we
have used the Tesseract training texts, which consists of about
170\,000 lines of random English `words' gleaned from a variety of
sources.  It is not clear how many lines of text it is best to train
on to avoid over-fitting, so for our experiments, we trained on 10\%
of them and also on 100\% of them; we compare the results below.  For
each of these, we randomly chose 90\% of the lines to be used for
training and 10\% for validation (`evaluation' in Tesseract's
terminology).  For the training on the entire 170\,000 text lines, we
performed this in 10 separate steps to allow the possibility of
evaluating the performance on different sizes of training data.  We
separately trained Tesseract for 60~dpi images and for 75~dpi images.

\begin{figure}[h]
  \begin{center}
    \includegraphics[width=0.8\linewidth]{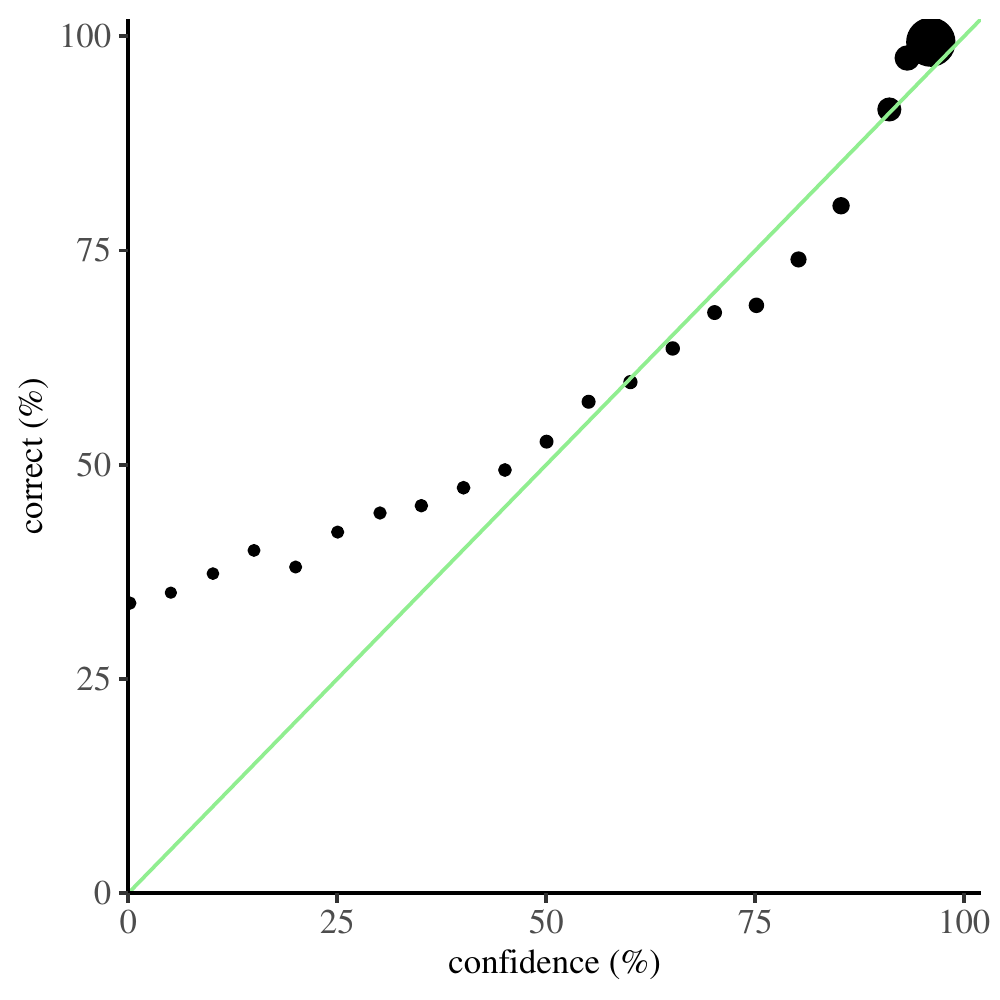}
  \end{center}
  \caption{The proportion of words in 20\,000 random lines of text
    (from the Tesseract training texts) that are correctly recognised,
    plotted against the confidence that Tesseract has in the word.
    This is the average across all seven upscaling methods produced
    for 60~dpi images.  The sizes of the dots are proportional to the
    frequency of words with this confidence, and words with confidence
    scores less than 95 are binned together.}
  \label{fig:confscores}
\end{figure}

\subsection{An ensemble method}

Initial small experiments showed that there was little consistency
about which type of upscaling gave the best recognition results across
texts; furthermore, this varied on a word-by-word basis.  We therefore
decided to use an ensemble method, combining the Tesseract outputs
from different upscalings to give the best possible output, at the
cost of running time.  As noted earlier, ensembling approaches to OCR
have been explored by a number of researchers.  However, we have more
information than they did: Tesseract can be instructed to provide a
confidence score and bounding box for each word.  (The confidence
score is determined from the final softmax layer and the CTC in a
somewhat complex way whose details do not concern us here.)  We
therefore wish to take account of the confidence of the different
outputs, not just their counts.  There are two key complexities to
address here:
\begin{enumerate}[(a)]
\item Determining when two words output by Tesseract represent the
  same piece of text; as can be seen in Figure~\ref{fig:merging}, the
  bounding boxes determined for the same word in different recognition
  passes can be very different, and one word can even appear as two
  words or be entirely missing in a different pass.
\item Deciding how to perform a majority vote taking confidences into
  account.
\end{enumerate}

There are several difficulties in implementing point~(a) above.  For
example, if a word appears at a particular location in one output but
no word appears at that location in another, what should we do?  If we
do decide to use that word, where in the text should it be inserted?
(We must recall that our primary goal is to recognise the original
text, and the order of words is a critical part of this, not just
their location on the page.)  This is further compounded when
Tesseract's page segmentation algorithm splits up the page in
different ways with different upscalings.  Several attempts to address
this were unsuccessful, and for this reason, we opted for a simpler
approach which works well most of the time.  We designate one output
as the master output, and compare all other outputs to that.  For
every word in the master output, we consider all the words in the
corresponding locations in the other outputs whose bounding box is
approximately the same as the master output's bounding box, and
ensemble the words so identified.  This means that the number of words
being ensembled at a particular point may be fewer than the number of
Tesseract outputs; in the example shown in Figure~\ref{fig:merging},
taking output~(a) as the master output, the final word
`Payment.\textquotedbl\,' does not match anything in output~(b), and
the words `Pa' and `l\textquotedbl\,' in output~(b) are ignored.
Finally, the choice of the master output was determined by the
performance of the modified Tesseract on a large sample of lines of
text drawn from the training set.  It turned out that for 60~dpi,
nearest neighbour plus a Gaussian filter with a standard deviation of
1~pixel was best, while bicubic interpolation was best for 75~dpi;
these were therefore used as the respective master outputs.

For point~(b), a further complication is the non-linearity of the
confidences in relation to the accuracy when working with
low-resolution images.  Figure~\ref{fig:confscores} shows the case for
a large set of 60~dpi images: while the word confidence scores for
300~dpi images turn out to be a fairly good estimate of the
probability of the word being correctly identified, this is not the
case for 60~dpi images.  There is also a lot of variation in this
between different fonts (not shown here).  We therefore work with a
\emph{modified confidence} score, which is a simple piecewise linear
function giving a very rough approximation to this data:
$$c_{\text{mod}}=\begin{cases}
  0.5c+30 & \text{if $c<80$} \\
  1.7c-65 & \text{if $c\ge80$}
  \end{cases}$$
(Though a confidence of 100 would give a modified confidence of over
100, Tesseract appears to never give a confidence score greater than
97.)  A similar function is used for 75~dpi images based on the
equivalent of Figure~\ref{fig:confscores}.

We now wish to perform a majority vote on each word, taking the
modified confidence scores into account.  There does not seem to be a
principled way to do this, as we do not know what probability each
network would give to other possible outputs.  So we resort to the
following: let the output words under consideration be $w^{(1)}$,
\dots, $w^{(n)}$ with respective modified confidences
$c_{\text{mod}}^{(1)}$, \dots, $c_{\text{mod}}^{(n)}$, and let $W$~be the
set of distinct output words.  Then we define the best output word to
be
$$w_{\text{best}} = \operatorname*{arg\,max}_{w\in W}
\frac{\sum_{\{i:w^{(i)}=w\}} c_{\text{mod}}^{(i)}}
{\sqrt{|\{i:w^{(i)}=w\}|}}$$
This is somewhere between the sum of all modified confidences for a
given possible output word and their mean, so that a repeated output
with modified confidences of~40 and~41 do not override a single output
with modified confidence of 80.  Prior to doing this, we also reduce
the modified confidence of outputs that are not dictionary words
(after removing any punctuation) by~30; this gives some extra weight
to dictionary words.  The choice of~30 was experimentally determined
as being a reasonable value on a small dataset, but slightly better
results might be obtained by using a different penalty.  We then
replace the master output word with the ensemble output to produce the
final output.

\section{Results and discussion}
\label{sec:results}

\subsection{Test data preparation}
\label{sec:data}

We tested our method on a large set of scanned images, none of which
had been used in the training or algorithm development process.
Furthermore, the training process generated images by simulation,
while the test data was produced using a real scanner.  To prepare the
images, we chose 11~different texts: ten were taken from a variety of
books in the Project Gutenberg~\cite{noauthor_project_nodate}
collection, including fiction, non-fiction and poetry, while one was
taken from a recent U.S.\ Supreme Court opinion.  (Details of the
texts used are included in the public dataset we have made available.)
Five pages of text were extracted from the start of each of these,
totalling 124\,961 characters and 25\,656 words.  Each page of text
was typeset in 18 different fonts (so identical text appeared on each
corresponding page).  These fonts included: serif and sans serif
fonts; monospaced and variable width fonts, and fonts which had been
trained on and fonts which had not been used for training.  The font
sizes were all 10~pt or 11~pt, with the font size chosen so that the
corresponding pages would each have about the same area of paper
covered with text.  The fonts used are listed in
Table~\ref{tab:fontsused}.  The resulting 990 pages were then printed
at 600~dpi and scanned at 300~dpi.  These scans were then down-sampled
to 60~dpi and 75~dpi using box interpolation (which takes the average
value of each $5\times5$ or $4\times4$ box of pixels respectively).

This variety allowed us to both test the effectiveness of our
algorithm and to independently assess its performance on different
types of text and on different fonts.

\begin{table}[th]
  \centering
  \begin{tabular}{lccccc}
    Font                  &Size&Serif&Mono&Train&Test\\ \hline
    Arial                 & 11 pt & no & no & yes & yes \\
    Avenir                & 11 pt & no & no & yes & no \\
    Baskerville           & 11 pt & yes & no & yes & yes \\
    Bookman Old Style     & 11 pt & yes & no & no & yes \\
    Calibri               & 11 pt & no & no & yes & yes \\
    Cambria               & 11 pt & yes & no & no & yes \\
    Century Gothic        & 10 pt & no & no & no & yes \\
    Chalkboard            & 11 pt & no & no & yes & no \\
    Comic Sans            & 10 pt & no & no & yes & yes \\
    Courier New           & 10 pt & yes & yes & yes & yes \\
    FrankRuehlCLM         & 11 pt & yes & no & yes & no \\
    Franklin Gothic Book  & 11 pt & no & no & yes & yes \\
    Futura                & 11 pt & no & no & yes & no \\
    Garamond              & 11 pt & yes & no & no & yes \\
    Gill Sans             & 11 pt & no & no & yes & yes \\
    Helvetica             & 11 pt & no & no & yes & yes \\
    Lucida Console        & 11 pt & no & yes & yes & no \\
    Lucida Sans           & 10 pt & no & no & yes & yes \\
    Menlo                 & 10 pt & no & yes & yes & yes \\
    Optima                & 11 pt & no & no & no & yes \\
    Palatino              & 11 pt & yes & no & yes & yes \\
    Tahoma                & 11 pt & no & no & yes & no \\
    Trebuchet             & 11 pt & no & no & yes & no \\
    Times New Roman       & 11 pt & yes & no & yes & yes \\
    Verdana               & 10 pt & no & no & yes & yes
  \end{tabular}
  \medskip
  \caption{Fonts used for training and testing}
  \label{tab:fontsused}
\end{table}


The character level accuracy (CLA) and word level accuracy (WLA) are
calculated by finding the Levenshtein distance~$d$ between our
algorithm's output and the ground truth; the percentage accuracy is
then calculated as $(1-\frac{d}{L})\times 100\%$, where $L$~is the
number of characters or words in the ground truth.  (For the CLA, we
first collapse all sequences of white space into a single space
character.)

One particular difficulty that quickly became apparent in initial
experiments is that it is almost impossible to identify the type of
quote marks used at these low resolutions; an open quote mark~(`), a
straight quote mark~(\textquotesingle) and a close quote mark~(') all
appear almost identical at 60~dpi.  So in calculating the CLA and WLA,
a decision was made to treat all three quote marks as equivalent (and
similarly for double quotes and for dashes/hyphens).

\subsection{Numerical results}

\begin{table*}[th]
  \centering
  \begin{tabular}{l|rrrr|rrrr|rr}
        &\multicolumn{2}{c}{60 dpi plain}&\multicolumn{2}{c|}{60 dpi lores}&
         \multicolumn{2}{c}{75 dpi plain}&\multicolumn{2}{c|}{75 dpi lores}&
         \multicolumn{2}{c}{300 dpi}\\
    Font&\multicolumn{1}{c}{CLA}&\multicolumn{1}{c}{WLA}&
         \multicolumn{1}{c}{CLA}&\multicolumn{1}{c|}{WLA}&
         \multicolumn{1}{c}{CLA}&\multicolumn{1}{c}{WLA}&
         \multicolumn{1}{c}{CLA}&\multicolumn{1}{c|}{WLA}&
         \multicolumn{1}{c}{CLA}&\multicolumn{1}{c}{WLA}\\ \hline
    Arial                & 99.4\% & 97.0\% & 99.8\% & 99.2\% & 99.8\% & 99.0\% &  99.9\% & 99.3\% &  99.9\% & 99.2\%\\
    Baskerville          & 99.1\% & 96.3\% & 99.6\% & 99.0\% & 99.8\% & 98.8\% &  99.8\% & 99.4\% &  99.9\% & 99.5\%\\
    Bookman Old Style    & 99.7\% & 98.3\% & 99.9\% & 99.7\% & 99.9\% & 99.5\% & 100.0\% & 99.8\% &  99.9\% & 99.6\%\\
    Calibri              & 99.4\% & 96.8\% & 99.8\% & 99.2\% & 99.8\% & 98.8\% &  99.9\% & 99.4\% &  99.9\% & 99.1\%\\
    Cambria              & 99.6\% & 97.8\% & 99.8\% & 99.0\% & 99.9\% & 99.1\% &  99.9\% & 99.7\% &  99.9\% & 99.4\%\\
    Century Gothic       & 96.6\% & 84.9\% & 99.2\% & 96.1\% & 99.2\% & 95.7\% &  99.7\% & 98.7\% &  99.7\% & 98.5\%\\
    Comic Sans           & 99.1\% & 95.1\% & 99.9\% & 99.4\% & 99.7\% & 98.3\% &  99.9\% & 99.6\% &  99.9\% & 99.3\%\\
    Courier New          & 98.9\% & 94.7\% & 99.8\% & 99.3\% & 99.7\% & 98.5\% &  99.9\% & 99.6\% &  99.9\% & 99.3\%\\
    Franklin Gothic Book & 99.1\% & 95.6\% & 99.8\% & 99.2\% & 99.7\% & 98.5\% &  99.9\% & 99.3\% &  99.9\% & 99.2\%\\
    Garamond             & 97.1\% & 91.2\% & 98.8\% & 97.6\% & 99.7\% & 98.6\% &  99.7\% & 99.2\% &  99.9\% & 99.3\%\\
    Gill Sans            & 99.0\% & 95.4\% & 99.7\% & 98.9\% & 99.8\% & 98.7\% &  99.8\% & 99.2\% &  99.9\% & 99.2\%\\
    Helvetica            & 99.1\% & 96.1\% & 99.8\% & 99.1\% & 99.8\% & 98.6\% &  99.8\% & 99.2\% &  99.8\% & 98.9\%\\
    Lucida Sans          & 99.5\% & 97.5\% & 99.8\% & 99.2\% & 99.8\% & 98.7\% &  99.9\% & 99.3\% &  99.9\% & 99.1\%\\
    Menlo                & 99.5\% & 97.3\% & 99.9\% & 99.6\% & 99.9\% & 99.3\% &  99.9\% & 99.7\% &  99.9\% & 99.6\%\\
    Optima               & 99.6\% & 97.9\% & 99.9\% & 99.3\% & 99.8\% & 99.0\% &  99.9\% & 99.3\% &  99.9\% & 99.3\%\\
    Palatino             & 99.0\% & 97.8\% & 98.5\% & 98.0\% & 99.8\% & 99.3\% &  99.7\% & 99.3\% & 100.0\% & 99.7\%\\
    Times New Roman      & 99.5\% & 97.4\% & 99.8\% & 99.4\% & 99.8\% & 99.2\% &  99.9\% & 99.7\% & 100.0\% & 99.7\%\\
    Verdana              & 99.5\% & 97.4\% & 99.9\% & 99.6\% & 99.9\% & 99.4\% &  99.9\% & 99.7\% & 100.0\% & 99.7\%\\
  \end{tabular}
  \medskip
  \caption{The average results on each font: the `plain' results are
    with standard Tesseract on images upscaled with bicubic
    interpolation, while the `lores' results are those obtained using
    our system.  The final two columns give the results for plain
    Tesseract on the original 300~dpi images.}
  \label{tab:results-byfont}
\end{table*}

We present results for the 18 fonts tested in
Table~\ref{tab:results-byfont}, averaged across all 55 pages printed
in that font.  The results are shown for the networks trained on 100\%
of the Tesseract training lines; the results when the networks were
trained on only 10\% of the lines are very slightly worse.  The table
also shows the performance of plain Tesseract on the images when they
have been upscaled to 300~dpi using bicubic interpolation.  Our
results show a significant improvement for 60~dpi images and a good
improvement for 75~dpi images (for 60~dpi, the overall character error
rate dropped by about 64\% and the word error rate by about 73\%; for
75~dpi, the corresponding figures are 35\% and 51\%).  The only font
which performs noticeably worse with our system is Palatino, and there
is no clear reason for this.

We note at this point that even for 60~dpi images, these results are
significantly better than the previous state-of-the-art's quoted
performance on 75~dpi images, and that the performance on both 60~dpi
and 75~dpi images is close to that of plain Tesseract on 300~dpi
images; for some fonts, our method even does slightly \emph{better} on
the 60~dpi images than plain Tesseract on the corresponding 300~dpi
images!

Inspired by this, we also considered what would happen if we started
with the 300~dpi scans and used our ensemble technique by taking the
plain Tesseract run as the master output, and the results of
downsampling the image to 75~dpi and using our modified Tesseract to
produce the other outputs.  Preliminary results indicate that the mean
WLA improves somewhat, but the CLA declines.  This is an area for
further study.

We next look at the impact of different texts on the performance of
this system.  We use Cambria to illustrate this, as its performance is
approximately in the middle of the fonts tested and it was not used
during the training.  Furthermore, we use the error rate (100\% minus
the accuracy rate) to make it easier to read the figures.

\begin{table}[th]
  \centering
  \begin{tabular}{lrrrr}
        &\multicolumn{2}{c}{60 dpi}&
         \multicolumn{2}{c}{75 dpi}\\
    Font&\multicolumn{1}{c}{C err}&\multicolumn{1}{c}{W err}&
         \multicolumn{1}{c}{C err}&\multicolumn{1}{c}{W err}\\ \hline
    Around the World     & 0.07\% & 0.35\% & 0.01\% & 0.08\% \\
    Best Poetry          & 0.33\% & 1.45\% & 0.20\% & 0.77\% \\
    David Copperfield    & 0.41\% & 1.84\% & 0.15\% & 0.82\% \\
    Engineering          & 0.06\% & 0.33\% & 0.01\% & 0.04\% \\
    English Church       & 0.11\% & 0.48\% & 0.01\% & 0.03\% \\
    Fire Prevention      & 0.13\% & 0.64\% & 0.02\% & 0.07\% \\
    Flatland             & 0.16\% & 0.71\% & 0.03\% & 0.15\% \\
    Practical Mechanics  & 0.34\% & 1.47\% & 0.12\% & 0.59\% \\
    Reflections          & 0.18\% & 0.89\% & 0.05\% & 0.20\% \\
    Supreme Court        & 0.38\% & 1.81\% & 0.22\% & 0.86\% \\
    Wordsworth           & 0.34\% & 1.75\% & 0.06\% & 0.44\%
  \end{tabular}
  \medskip
  \caption{The performance of our system on a variety of texts printed
    in Cambria}
  \label{tab:results-texts}
\end{table}

It is pleasing to see that the performance was fairly consistent
across these different text genres, ranging from 19th century poetry
(Best Poetry and Wordsworth), through 19th century novels (Around the
World in Eighty Days, for example) and non-fiction works (Fire
Prevention, for example), to a recent opinion of the United States
Supreme Court.  There is no obvious pattern to which texts which
performed better; one might have anticipated poetry doing very poorly,
but that has not turned out to be the case.  We might have achieved
better results for specific texts if we had fine-tuned the network
using related lines of text (as Wang and
Singh~\cite{wang_systems_2018} did), as the LSTM would have learnt to
recognise word patterns, but then it might have performed more poorly
on other texts.  This remains a possible way to improve our system.

\section{Conclusion}
\label{sec:conclusion}

We have presented a novel end-to-end ensembling approach for
performing optical character recognition on very-low-resolution
scanned documents.  We have made use of what is known about human
perception of letters in order to improve our training approach over
the existing state-of-the-art, and demonstrated that our approach has
very high average accuracy levels on a large set of images.  We were
able to successfully perform OCR on lower resolution images than
previously reported, achieving character and word level accuracies of
over 99\% for most fonts scanned at 60~dpi, and also to achieve an
accuracy rate on 75~dpi scans similar to that on the equivalent
300~dpi images, far exceeding the existing state-of-the-art
performance.  We noted that different fonts give different results
with our system, and highlighted the need for a standard benchmark
dataset for this type of problem.

There are many open questions remaining.  On the technical side, what
is the best number of training iterations to perform?  How could we
improve the ensembling algorithm?  An improvement would be to perform
the page segmentation using just one upscaling, and to use this
segmentation for each of the line recognition runs; we did not attempt
to implement this in our experiments.  We could also use fewer
upscalings or different upscalings from the ones we chose: it is
unclear how to best choose them, and our choice was a somewhat
arbitrary selection of filters for reducing the high frequency
components of the images.

Finally, it would be very useful to adapt this approach to tackle a
challenge such as the ICDAR 2015 Competition on Text Image
Super-Resolution~\cite{peyrard_icdar2015_2015}, which used very
different types of image from the ones we studied in this work.

\section*{Software and dataset availability}

The modified version of Tesseract 4.1.1 used in these experiments can
be found at
\url{https://github.com/juliangilbey/tesseract/tree/lores-v1.0}.
(This URL points to the exact version used in these experiments.)
Supporting scripts used are available are
\url{https://github.com/juliangilbey/ocr-lowresolution}.  The scanned
image dataset is available at
\url{http://doi.org/10.5281/zenodo.3945525} and the trained Tesseract
networks are available from
\url{http://doi.org/10.5281/zenodo.3945949}.

\section*{Acknowledgements}

We thank Cognizant, in particular Indranil Mitra, Aditya Joshi and
Shishir Bharti, for posing this industrial problem and for funding
this work through research grant RG299289.  We also thank Linda Bowns
for pointing us in the direction of spatial frequency channels, which
led us to Majaj et al's work~\cite{majaj_role_2002}.

{\small
\bibliographystyle{IEEEtrandoi}
\bibliography{loresbib}
}

\end{document}